\documentclass[conference]{IEEEtran}
\IEEEoverridecommandlockouts
\usepackage{cite}
\usepackage{amsmath,amssymb,amsfonts}
\usepackage{algorithmic}
\usepackage{graphicx}
\usepackage{textcomp}
\usepackage{xcolor}
\def\BibTeX{{\rm B\kern-.05em{\sc i\kern-.025em b}\kern-.08em
    T\kern-.1667em\lower.7ex\hbox{E}\kern-.125emX}}
\usepackage[hyphens]{url}
\usepackage[normalem]{ulem}
\usepackage{hyperref}
\usepackage[hyphenbreaks]{breakurl}
\usepackage{ascmac}
\usepackage{tcolorbox}
\usepackage{booktabs}
\tcbuselibrary{raster,skins,breakable}
\tcbuselibrary{xparse} 
\DeclareTColorBox{brekableitembox}{ o m O{.5} O{} }%
  {empty, left=2mm, right=2mm, top=-2mm, bottom=0.2mm, attach boxed title to top left={xshift=1.2em},
  boxed title style={empty,left=-2mm,right=-2mm}, colframe=black, coltitle=black, coltext=black, breakable,  
  underlay unbroken={\draw[black,line width=#3pt]
    (title.east) -- (title.east-|frame.east) -- (frame.south east) -- (frame.south west) -- (title.west-|frame.west) -- (title.west); },
  underlay first={\draw[black,line width=#3pt](title.east) -- (title.east-|frame.east) -- (frame.south east) ;
    \draw[black,line width=#3pt] (frame.south west) -- (title.west-|frame.west) -- (title.west); },
  underlay middle={\draw[black,line width=#3pt](frame.north east) -- (frame.south east) ;
    \draw[black,line width=#3pt](frame.south west) -- (frame.north west) ;},
  underlay last={\draw[black,line width=#3pt](frame.north east) -- (frame.south east) -- (frame.south west) -- (frame.north west) ;}, IfValueTF={#1}{title=~~#2~~〈#1〉}{title=~~#2~~},#4}

\begin{document}

\title{From Base to Conversational: Japanese Instruction Dataset and Tuning Large Language Models\\

}

\author{\IEEEauthorblockN{Masahiro Suzuki}
\IEEEauthorblockA{
\textit{The University of Tokyo}\\
Tokyo, Japan \\
research@msuzuki.me
}\and
\IEEEauthorblockN{Masanori Hirano}
\IEEEauthorblockA{
\textit{The University of Tokyo}\\
Tokyo, Japan \\
research@mhirano.jp}
\and
\IEEEauthorblockN{Hiroki Sakaji}
\IEEEauthorblockA{
\textit{The University of Tokyo}\\
Tokyo, Japan \\
sakaji@sys.t.u-tokyo.ac.jp
}
}

\maketitle

\begin{abstract}
Instruction tuning is essential for large language models (LLMs) to become interactive.
While many instruction tuning datasets exist in English, there is a noticeable lack in other languages.
Also, their effectiveness has not been well verified in non-English languages.
We construct a Japanese instruction dataset by expanding and filtering existing datasets and apply the dataset to a Japanese pre-trained base model.
We performed Low-Rank Adaptation (LoRA) tuning on both Japanese and English existing models using our instruction dataset.
We evaluated these models from both quantitative and qualitative perspectives.
As a result, the effectiveness of Japanese instruction datasets is confirmed.
The results also indicate that even with relatively small LLMs, performances in downstream tasks would be improved through instruction tuning.
Our instruction dataset, tuned models, and implementation are publicly available online.
\end{abstract}

\begin{IEEEkeywords}
Large Language Model (LLM), Instruction Dataset, Instruction Tuning, Japanese
\end{IEEEkeywords}

\section{Introduction}
Large language models (LLMs) have been making remarkable progress in performance and generalization in recent years.
Various Transformer-based~\cite{Vaswani2017} language models, such as BERT~\cite{Devlin2018}, RoBERTa~\cite{Liu2019roberta}, and the GPT series~\cite{GPT-1,GPT-2,GPT-3}, have demonstrated high performance derived from pre-training.
Furthermore, since 2022, a large number of models, such as OPT~\cite{zhang2022opt}, GPT-NeoX-20B~\cite{black-etal-2022-gptneox}, UL2~\cite{tay2023ul}, PaLM~\cite{chowdhery2022palm}, BLOOM~\cite{scao2022bloom}, Pythia~\cite{biderman2023pythia}, and LLaMA series~\cite{touvron2023llama,touvron2023llama2}, have emerged as models that show higher performance by scaling their size~\cite{wei2022emergent}.

Although there is still difficulty in few-shot or zero-shot performance on unseen tasks, instruction tuning can address this issue~\cite{wei2022flan}.
Instruction tuning is a training method that improves the performance in unseen tasks by solving various tasks described via natural language instructions~\cite{wei2022flan}.
Starting with the enhancement of performance in various tasks by GPT-3~\cite{GPT-3} under a few-shot setting given in natural language, there has been an increasing demand for responses in formats that are closer to question-answering or conversation, especially formats that are not similar to the pre-training data.

An increasing number of datasets for instruction tuning and instruct-tuned models are being made available to the public. 
For instance, various datasets like FLAN~\cite{wei2022flan}, P3~\cite{sanh2022multitaskp3}, databricks-dolly-15k~\footnote{\url{https://huggingface.co/datasets/databricks/databricks-dolly-15k}}, and OASST1~\cite{köpf2023openassistant} have been proposed and made public.
As publicly available models, Flan-T5~\cite{chung2022scaling} was constructed using FLAN and T0 was constructed using P3 respectively.
Also, Dolly~\cite{dolly} is a model with instruction tuning applied to Pythia~\cite{biderman2023pythia}, while Vicuna~\cite{vicuna} and Alpaca~\cite{alpaca} are models with instruction tuning applied to LLaMA~\cite{touvron2023llama}.

However, these models are not fully compatible with languages other than English.
The datasets used for instruction tuning in Dolly, Alpaca, and Vicuna are only in English, making it difficult to gain the benefits of these models in languages other than English.
Many instruction datasets have been constructed in English, and there are not many efforts to construct instruction datasets in languages other than English.
While there are movements to construct instruction datasets in Chinese~\cite{zhang2023chinese}, most instruction dataset in non-English languages are built from outputs of models with licensing restrictions, such as translations of the Alpaca dataset~\cite{alpaca} or the ShareGPT52K~\footnote{\url{https://huggingface.co/datasets/RyokoAI/ShareGPT52K}} constructed from ChatGPT outputs.
In languages other than English, the scarcity of comprehensive instruction datasets means that the verification of instruction tuning effects is limited~\cite{cui2023efficient}.
In Japanese, only some data from translated Alpaca~\cite{alpaca} and OASST1~\cite{köpf2023openassistant} exists, and there's a lack of dataset diversity, with quantitative evaluations of instruction tuning yet to be conducted.
While constructing and evaluating datasets in languages other than English is a crucial step towards building language models that can interact in various languages, it's still very much in its early stages.

To tackle the issue of the lack of Japanese instruction dataset, the study~\cite{hirano2023llmjapanesedataset} gathers various Japanese datasets to build an instruction dataset.
While this dataset seems valuable, the effect of instruction tuning is only shown qualitatively and not quantitatively.
Furthermore, the majority of this dataset consists of translation tasks.
While it is considered that the translation tasks are effective when adapting English-based models to Japanese, these tasks may not be optimal for Japanese-based models.
To apply the instruction dataset to a Japanese-based model, it is desirable to filter out the translation data and construct an instruction dataset consisting solely of Japanese.

We construct an instruction dataset consisting solely of Japanese for instruction tuning based on a Japanese model by filtering and expanding the Japanese instruction dataset~\cite{hirano2023llmjapanesedataset}.
The constructed dataset contains about 2.5 million samples and 5 tasks, such as commonsense, summarization, reading comprehension, simplification, and correction.
Using this dataset, which contains various tasks, we perform instruction tuning on both Japanese-based and English-based LLMs.
For Japanese-based models, we conduct tuning using an instruction dataset without translation data, while for English-based models, we do using an instruction dataset that includes translation data.
As a result of quantitative evaluation with the tuned model, we demonstrate that instruction tuning in Japanese improve the performance in downstream tasks, thereby illustrating the effectiveness of the Japanese instruction dataset.
The following materials used in this study are available as open source.
\begin{itemize}
    \item Japanese instruction dataset: \\ \url{https://huggingface.co/datasets/izumi-lab/llm-japanese-dataset-vanilla}
    \item Tuned model (Stormy 10 epochs): \\ \url{https://huggingface.co/izumi-lab/stormy-7b-10ep}
    \item Tuned model (LLaMA 7B 5 epochs): \\ \url{https://huggingface.co/izumi-lab/llama-7b-japanese-lora-v0-5ep}
    \item Implementation for training and evaluation: \\ \url{https://github.com/retarfi/jallm}
\end{itemize}
Here are our main contributions: (1) We construct a Japanese instruction dataset, llm-japanese-dataset-vanilla, for Japanese-based models; (2) We clarified the benefits of instruction tuning for Japanese and English models from evaluating with some Japanese downstream tasks; (3) Unlike previous research~\cite{wei2022flan}, we show that even with smaller model sizes, instruction tuning can lead to performance gains in downstream tasks.

\begin{figure}[tb]
  \centering
  \includegraphics[width=\columnwidth, keepaspectratio=true]{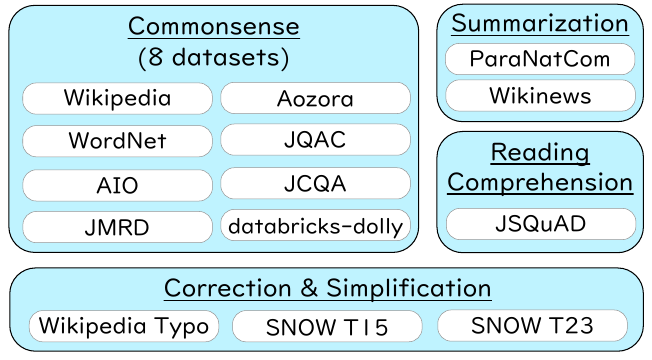}
  \caption{Datasets and task clusters used in llm-japanese-dataset-vanilla v1.0.1.}\label{fig:data-consists}
\end{figure}

\section{Instruction Dataset}
We construct a Japanese instruction dataset without translation tasks.
We use the llm-japanese-dataset v0.1.0~\cite{hirano2023llmjapanesedataset} as a main data source for the Japanese instruction dataset and expand this dataset with additional Japanese datasets.
The llm-japanese-dataset v0.1.0 contains about 8.4 million instruction examples, of which more than 75 \% (6,581,044) are constructed based on translation data.
This dataset is intended to link English and Japanese and extract the knowledge learned in English for use in Japanese as well, considering that many LLMs like LLaMA show good performance in English.
However, when it comes to Japanese-based models, they are usually pre-trained with Japanese corpora.
The need for the English part of this dataset is relatively low because the part aimed to link English and Japanese.
Therefore, we extract 1,811,964 data excluding translation tasks from the llm-japanese-dataset v0.1.0.
Furthermore, to expand the variety of datasets, we incorporated the Japanese Wikipedia Typo Dataset (Wikipedia Typo)~\cite{tanaka2021typo} and the Japanese Question-Answering Corpus (JQAC)~\cite{tanioka2018jqac}.
From the Wikipedia Typo and JQAC, we newly created 697,316 and 906 instruction entries respectively.
Additionally, we addressed licensing issues present in version v0.1.0, and ultimately constructed a total of 2,463,624 instruction data entries, releasing it as llm-japanese-dataset-vanilla v1.0.1~\footnote{\url{https://huggingface.co/datasets/izumi-lab/llm-japanese-dataset-vanilla}}.
Figure \ref{fig:data-consists} shows datasets and task classifications included in llm-japanese-dataset-vanilla v1.0.1.

We use the instruction, input, and response included in llm-japanese-dataset-vanilla v0.1.0, following the format below.

\setcounter{footnote}{2}
\newcounter{footnotecommon1}
\footnotetext{Originally written in Japanese.}\setcounter{footnotecommon1}{\value{footnote}}
\vspace{-2mm}\begin{brekableitembox}{Prompt format with input}
\fontsize{9.5pt}{13pt}\selectfont\uline{Below is an instruction that describes a task, paired with an input that provides further context. Write a response that appropriately completes the request.
\\\\
\#\#\# Instruction:\\
\{Instruction\}
\\\\
\#\#\# Input:\\
\{Input\}
\\\\
\#\#\# Response:\\
\{Response\}
}~\footnotemark[\value{footnotecommon1}]
\end{brekableitembox}

\vspace{-2mm}\begin{brekableitembox}{Prompt format with no input}
\fontsize{9.5pt}{13pt}\selectfont\uline{Below is an instruction that describes a task. Write a response that appropriately completes the request.
\\\\
\#\#\# Instruction:\\
\{Instruction\}
\\\\
\#\#\# Response:\\
\{Response\}
}~\footnotemark[\value{footnotecommon1}]
\end{brekableitembox}

\begin{table*}[tb]
\caption{The parameters of LoRA tuning}
\label{tab:tune-params}
\begin{center}
\begin{tabular}{cccc} \toprule
  Parameter & Stormy & Instruct LLaMA 7B & Instruct LLaMA 13B\cite{hirano2023llmjapanesedataset} \\ \midrule
  Base Model & CALM 7B & LLaMA 7B & LLaMA 13B \\
  Learning Rate & 3e-4 & 3e-4 & 3e-4 \\
  Sequence Length & 300 & 256 & 256 \\
  Batch & 128 & 128 & 130 \\
  \# of data & 1.4M & 8.4M & 8.4M \\
  Epochs & 10 & 5 & 1 \\
  $r$ in LoRA & 4 & 4 & 4 \\
  $\alpha$ in LoRA & 16 & 16 & 16 \\
  Dropout Ratio of LoRA & 0.05 & 0.05 & 0.05 \\
  Tuning Parameters & query\_key\_value & q\_proj, v\_proj & q\_proj, v\_proj \\
  \bottomrule
\end{tabular}
\end{center}
\end{table*}

\section{Instruction LoRA Tuning}
We perform Low-Rank Adaptation (LoRA) tuning~\cite{hu2022lora} on two publicly available LLMs.
In this section, we describe the base model and the process of LoRA tuning.

\subsection{Models}
We use two models: a Japanese-based model and an English-based model.
The models we use are the Japanese-based OpenCALM-7B (hereafter CALM) and the English-based LLaMA 7B.
CALM is a model with 7 billion parameters released by CyberAgent~\footnote{\url{https://huggingface.co/cyberagent/open-calm-7b}}.
It is pre-trained on Japanese Wikipedia and Common Crawl using the GPT-NeoX architecture~\cite{black-etal-2022-gptneox}.
For the English-based model, we use the 7B model of LLaMA~\cite{touvron2023llama} (hereafter LLaMA 7B), which is released by Meta~\footnote{Strictly speaking, although it was not initially open-source, it has become available under certain licenses}.
Although LLaMA is trained in English and is not specialized for Japanese, it is capable of Japanese input and output.
Even for LLaMA, we attempt to output in Japanese by conducting instruction tuning and evaluation experiments using Japanese contexts.

Due to the differences in characteristics between the Japanese-based CALM and the English-based LLaMA 7B, we use llm-japanese-dataset-vanilla, which we constructed above, for CALM and llm-japanese-dataset for LLaMA 7B as training data.
For tuning CALM, we use version v0.1.0 as the training data, which excludes the JQAC and Wikipedia Typo datasets.
This is done to align with the model constructed in the literature~\cite{hirano2023llmjapanesedataset}, ensuring dataset consistency with the exception of not including English.

We train LLaMA 7B on the entire llm-japanese-dataset v0.1.0, following the methods outlined in~\cite{hirano2023llmjapanesedataset}.
We adopt the same input format as described in~\cite{hirano2023llmjapanesedataset}.
From this point forward, the tuned CALM will be referred to as ``Stormy,'' and the LLaMA 7B as ``Instruct LLaMA 7B.''

\subsection{LoRA Tuning}
LLMs, having a large number of parameters, require GPU resources not only for pre-training but also for fine-tuning.
In this study, we use LoRA~\cite{hu2022lora} as a method for tuning LLMs without significantly reducing accuracy.
In LoRA, only the difference between the initial and updated LLM parameters, represented with small-scale parameters, is calculated.
Consider an example of updating the parameter matrix $W_0 \in \mathbb{R}^{d\times k}$ of a certain linear layer that LLM has.
Instead of training $W_0$ directly, initialize the difference $\Delta W \in \mathbb{R}^{d\times k}$ to $W_0$ with a zero matrix, update the difference $\delta W$, and proceed with training by updating the parameters to $W_0 + \Delta W$.
Here, we set $\Delta W= BA$ where $B \in \mathbb{R}^{d \times r}$ and $A \in \mathbb{R}^{r \times k}$ are matrices of rank $r \ll \min (d,k)$.
This can reduce the number of learnable parameters from $dk$ to $(d+k)r$.

The primary parameters utilized in the experiment are shown in Table \ref{tab:tune-params}.
For comparison, we also mention the model that Instruct LLaMA 13B~\cite{hirano2023llmjapanesedataset}, which was LoRA-tuned with llm-japanese-dataset v0.1.0.

We used PEFT~\cite{PEFT} and DeepSpeed ZeRO~\cite{rajbhandari2020zero} for implementation.
The code is available at \url{https://github.com/retarfi/jallm}.

\section{Evaluating Constructed Models}
We evaluate the tuned models both quantitatively and qualitatively.
From the quantitative perspective, we evaluate from two perspectives.
The first is accuracy derived from the likelihood of choices in text classification tasks with JNLI and MARC-ja.
JNLI and MARC-ja are tasks from JGLUE~\cite{kurihara-etal-2022-jglue}.
Further details are described in Section \ref{sec:eval-accuracy}.
The second is perplexity using question-answering data that is not included in the dataset constructed in this study.
From the qualitative perspective, we qualitatively evaluate the output for several prompts.
The temperature parameter for generation is 0.0, and the repetition penalty~\cite{nitish2019repetition} is 1.05 for CALM and Stormy and 1.0 for Instruct LLaMA 7B and LLaMA 7B.
We use 5 prompts for input to the models, which are the same as those used in the literature~\cite{hirano2023llmjapanesedataset}.
We also conduct evaluation experiments on LLaMA 13B and Instruct LLaMA 13B, which  was instruction tuned for LLaMA 13B, constructed in the study~\cite{hirano2023llmjapanesedataset} as well.

\subsection{Accuracy}\label{sec:eval-accuracy}
Another evaluation is performed by JNLI and MARC-ja included in JGLUE~\cite{kurihara-etal-2022-jglue}.
JNLI is a task to choose the relationship that the premise sentence shows to the sentence pair of the hypothesis from three options: entailment, contradiction, and neutral.
MARC-ja is a task to choose either ``positive'' or ``negative'' in Japanese for product reviews and is constructed using the Japanese part of the Multilingual Amazon Reviews Corpus (MARC) ~\cite{marc_reviews}.
In addition to these, JGLUE includes JCommonsenseQA, which questions common sense, and JSQuAD, which is an extraction task.
However, these data are included in the llm-japanese-dataset v0.1.0, which is used for instruction LoRA tuning.
Therefore, they were considered inappropriate as evaluation tasks and excluded.

For the implementation of the experiment, we use the Japanese evaluation branch~\footnote{\url{https://github.com/Stability-AI/lm-evaluation-harness/tree/jp-stable}} of Stability-AI/lm-evaluation-harness~\cite{eval-harness}.
Aligning with lm-evaluation-harness, we use the prompt version that achieves the best performance.
We adopt v0.2 for Stormy and v0.3 for the others, such as CALM, Instruct LLaMA 7B, LLaMA 7B, Instruct LLaMA 13B, and LLaMA 13B.
Detailed prompts are described in the \hyperref[sec:prompt-jnli-marcja]{Appendix}.

For the input prompt, we compare the likelihood of the strings of each task's choicesand take the highest one as the model's output.
In JNLI, the three choices are entailment, contradiction, and neutral, and in MARC-ja, the two choices are ``positive'' and ``negative'' in Japanese, and the model outputs the choice with the highest likelihood of output.
Therefore, outputs other than the choices are not considered.
We evaluate for each of 1-shot, 2-shot, and 3-shot, which show one, two, or three examples in the input, respectively.

\begin{table*}[htb]
\centering
\caption{Results of the evaluation experiment. $^*$ indicates that there were data in the evaluation dataset that exceeded the input length of LoRA tuning (Stormy is 300, Instruct LLaMA 7B and Instruct LLaMA 13B are 256). $^{\dagger}$ indicates that there were data in the evaluation dataset that exceeded the maximum input length of the model (both CALM-based and LLaMA-based are 2,048). The highest-performing areas for each task are indicated in bold.}
\label{tab:eval-result}
\begin{tabular}{cccccccc} \toprule
   & \multicolumn{3}{c}{JNLI} & \multicolumn{3}{c}{MARC-ja} & VQA \\ \cmidrule(lr){2-4} \cmidrule(lr){5-7}
  Model & 1-shot & 2-shot & 3-shot & 1-shot & 2-shot & 3-shot & Perplexity \\ \midrule
  Stormy (Instruct CALM) & \bf{0.459} & \bf{0.508} & 0.475 & 0.468 & 0.828$^*$ & 0.784$^*$ & \bf{29.9} \\
  CALM\footnotemark[1] & 0.294 & 0.331 & 0.314 & 0.781 & 0.836 & \bf{0.856} & 246.6 \\
  Instruct LLaMA 7B & 0.398$^*$ & 0.454$^*$ & \bf{0.479}$^{*\dagger}$ & 0.795$^*$ & 0.829$^*$ & 0.847$^*$ & 68.5 \\
  LLaMA 7B\cite{touvron2023llama} & 0.171 & 0.273 & 0.303$^{\dagger}$ & 0.839 & 0.848 & 0.852 & 1,499 \\
  Instruct LLaMA 13B~\cite{hirano2023llmjapanesedataset}& 0.302$^*$ & 0.302$^*$ & 0.302$^{*\dagger}$ & \bf{0.859}$^*$ & \bf{0.855}$^*$ & 0.855$^*$ & 38.8\\
  LLaMA 13B~\cite{touvron2023llama} & 0.316 & 0.281 & 0.263$^{\dagger}$ & 0.855 & \bf{0.855} & 0.855 & 971.5 \\
  \bottomrule
\end{tabular}
\end{table*}

\subsection{Perplexity}
Perplexity, as defined by ~\cite{jelinek1977perplexity}, is the exponential of the average negative log-likelihood.
The lower the value, the higher the probability that the words in the dataset are correctly output.
Given a tokenized sequence $X=(x_0, x_1, \cdots, x_t)$, the perplexity of $X$ is represented by Equation \eqref{eq:ppl}.

\begin{equation}\label{eq:ppl}
  \mathrm{Perplexity}(X) = \exp \left\{-\frac{1}{t}\sum_{i}^{t} \log p_{\theta}(x_i|x_{<i})\right\}
\end{equation}

Here, $\log p_{\theta}(x_i|x_{<i})$ is the log-likelihood of the $i$-th token given the preceding tokens $x_{<i}$.

In this study, we measure perplexity using the Japanese Visual Question Answering (VQA) dataset~\cite{shimizu2018vqa}, which is not included in the llm-japanese-dataset v0.1.0 used for tuning the language model.
Although this VQA dataset is a question-answering task performed by looking at presented images, it is conjectured that models with a high probability of predicting the correct response sentence are more natural.
We convert 793,664 question and answer pairs extracted from the VQA dataset into prompt format and input them.
An example of the input is shown below.

\vspace{-2mm}\begin{brekableitembox}{Example in VQA with Japanese-based Model}
\fontsize{9.5pt}{13pt}\selectfont\uline{Write a response to answer the following question.
\\\\
\#\#\# Question:\\
What color is the airplane's body?
\\\\
\#\#\# Response:\\
White}~\footnotemark[\value{footnotecommon1}]
\end{brekableitembox}

It should be noted that the LLaMA-based model uses English for system messages and Japanese for contexts of questions andw responses.
Therefore, following the literature~\cite{hirano2023llmjapanesedataset}, the above example is modified as follows.

\vspace{-2mm}\begin{brekableitembox}{Example in VQA with English-based Model}
\fontsize{9.5pt}{13pt}\selectfont Write a response to answer the following question.
\\\\
\#\#\# Question:\\
\uline{What color is the airplane's body?}~\footnotemark[\value{footnotecommon1}]
\\\\
\#\#\# Response:\\
\uline{White}~\footnotemark[\value{footnotecommon1}]
\end{brekableitembox}

The calculation of perplexity is not performed on the input to the model and is only applied to the response. 
In other words, in the above example, perplexity is calculated only for the token corresponding to the output ``white.''

\section{Results and Discussion}
\subsection{Quantitative Evaluation}\label{sec:quantitative-eval}
Table \ref{tab:eval-result} shows the results of the evaluation experiments.

In the evaluation by JNLI, the accuracy of Stormy was the highest across 1-shot, 2-shot, and 3-shot settings.
Even though the llm-japanese-dataset v0.1.0 does not include a dataset equivalent to implication relation recognition, the performance seems to have been improved by solving various tasks as in \cite{wei2022flan}.
The improvement in performance on tasks not present in the dataset by using instruction tuning across various tasks aligns with the findings in the literature~\cite{wei2022flan,sanh2022multitaskp3}.
Japanese instruction datasets are valuable in the point of having constructed similar datasets for languages other than English.
The performance of Stormy and Instruct LLaMA 7B, which performed instruction tuning on CALM and LLaMA 7B, respectively, improved, showing the effect as instruction tuning.
However, the effect of instruction tuning in LLaMA 13B was relatively small.
This is likely because instruction tuning in Instruct LLaMA 13B was performed for only one epoch.
When comparing two Instruct LLaMA models with different numbers of parameters, even though there was a difference in the number of training epochs, Instruct LLaMA 7B showed a stronger effect from instruction tuning.
This is considered to be because the smaller model size facilitates more effective training.
It has been reported that larger model sizes result in better performance on downstream tasks~\cite{zhang2022opt,chowdhery2022palm,touvron2023llama}.
The performance of Instruct LLaMA 13B might improve with more training epochs.

In the evaluation by MARC-ja, there was no performance improvement by instruction tuning in all of 1-shot, 2-shot, and 3-shot, or the performance became worse.
This phenomenon has also been reported in \cite{ouyang2022instructgpt,wei2022flan}.
The performance might be improved by adopting more various tasks widely as instruction data as in  \cite{wei2022flan}.
As well as MARC-ja, there are also datasets related to sentiment that can be incorporated in Japanese, such as the chABSA-dataset\footnote{\url{https://github.com/chakki-works/chABSA-dataset}} (ABSA stands for Aspect-Based Sentiment Analysis).
The decrease in accuracy could be suppressed by additionally training these datasets.
Another possible reason why the performance did not improve in the LLaMA-based models is the input length of instruction tuning in this study.
While the LLaMA-based model itself can input up to 2,048 tokens and pre-training is performed at this length, in this study, the input length is limited to 256 tokens.
Therefore, in data where long tokens are input, the effect of instruction tuning may not have been demonstrated.
Extending the input length of instruction tuning is a future issue.

In the evaluation of perplexity using VQA, all the instruct-tuned models showed improved performance with reduced perplexity due to tuning using instruction data.
Language models adopting the decoder architecture are trained to increase the probability of correctly predicting the next token in the input.
Therefore perplexity is trained to decrease.
However, the reason for the reduction in perplexity by instruction tuning might be attributed to differences in the input data.
While a language model predicts the next token for consecutive sequences in pre-training, it predicts tokens sequentially in response to a given question in instruction tuning.
Since the format of input and output in instruction tuning matches the question-answering in VQA used in this experiment, it can be inferred that the model became more accustomed to producing answers by instruction tuning, leading to a reduction (performance improvement) in perplexity.

The improvement in perplexity was particularly noticeable in the LLaMA-based models.
A link with Japanese is considered to have been created and the performance improved by training using instruction data including translation data, even for models other than Japanese, such as English.
Among the six models, the one with the highest perplexity and the worst performance was LLaMA 7B.
This is thought to be due to the fact that it is an English-based model and has fewer model parameters than LLaMA 13B.
On the other hand, the model that showed the best performance with the lowest perplexity was Stormy.
The performance was improved by further instruction tuning for CALM, which is a model of Japanese.
Comparing CALM, LLaMA 7B, and LLaMA 13B, which were the base models for tuning, the Japanese-based CALM showed the highest performance.

In terms of the effect of instruction tuning from the perspective of model size, the literature~\cite{wei2022flan} reported that for models larger than 68B, the effects of instruction tuning were observed in downstream tasks.
However, they also reported for models smaller than 8B, instruction tuning paradoxically degraded performance in downstream tasks.
In the results of the MARC-ja experiments in this study, no effect of instruction tuning was observed for all models of 7B and 13B, while for JNLI, the positive effects of instruction tuning were observed in all models.
This effect was observed in both Japanese-based CALM and English-based LLaMA models.
This suggests that, in non-English languages or when tuning English models to them, instruction tuning does not necessarily have negative effects for smaller models, and could even contribute to performance enhancement.

in comparison with prior research~\cite{wei2022flan,wang-etal-2022-super}, there are fewer types of tasks.
This might have led to a potential constraint in performance improvement.
For instance, when compared to FLAN~\cite{wei2022flan}, tasks like simplification and correction have been newly added, while tasks like natural language inference, sentiment, and paraphrase lack.
In this study, although the experiments were conducted using the 5 task types shown in Figure \ref{fig:data-consists}, consistent results were observed even in a non-English language like Japanese.
Expanding the variety of tasks will be a challenge for future research.

Regarding the base language of the model, there was no difference in terms of performance trends between the Japanese-based model and the English-based model; performance improved in VQA and JNLI, but not in MARC-ja.
The results show that using a Japanese instruction dataset can enhance performance in Japanese tasks for both English-based and Japanese-based models.
Except for the fact that translation tasks were used in the instruction tuning data of the LLaMA-derived model, the other tasks used in Stormy and Instruct LLaMA are the same.
How the translation task impacts both English-based and Japanese-based models remains a future research topic.

As discussed in the results for JNLI, while instruction tuning is effective for unseen tasks, the inclusion of similar tasks in the dataset is considered to have a substantial impact on performance improvement~\cite{wei2022flan}.
Based on this, it can be inferred that by adding tasks related to natural language inference for JNLI and sentiment for MARC-ja, the performance on JNLI and MARC-ja tasks would improve more.
The tasks evaluated in this study evaluate how much the model can output for a certain output (although it is consistent with the evaluation method in \cite{GPT-3}), and do not consider the actual output of the model.
For a more detailed comparison, it is also a future task to evaluate using actual outputs.
In addition, it will be necessary in the future to evaluate from the perspective of human preference~\cite{peng2023instruction,zhou2023lima}, not just focusing on the task's correctness.

\subsection{Qualitative Evaluation}
The following shows the responses of the models used in this study, such as Stormy, CALM, Instruct LLaMA 7B, and LLaMA 7B.
For responses of Instruct LLaMA 13B and LLaMA 13B to the input shown below, please refer to the literature~\cite{hirano2023llmjapanesedataset}.

\vspace{-2mm}\begin{brekableitembox}{Comparison of Responses (1)}
\fontsize{9.5pt}{13pt}\selectfont\textbf{\#\#\# Input:}\\
\uline{Who are you?}~\footnotemark[\value{footnotecommon1}]\\
\textbf{\#\#\# Output (Stormy):}\\
\uline{I am my wife, and I love her.}~\footnotemark[\value{footnotecommon1}]\\
\textbf{\#\#\# Output (CALM):}\\
\uline{I am Nick M. Kelemel. I have degrees in computer science and mathematics, and I am currently working as a researcher.}~\footnotemark[\value{footnotecommon1}]\\
\textbf{\#\#\# Output (Instruct LLaMA 7B):}\\
\uline{What are you doing?}~\footnotemark[\value{footnotecommon1}]\\
\textbf{\#\#\# Output (LLaMA 7B):}\\
\uline{Who are you?}~\footnotemark[\value{footnotecommon1}]
\end{brekableitembox}

\vspace{-2mm}\begin{brekableitembox}{Comparison of Responses (2)}
\fontsize{9.5pt}{13pt}\selectfont\textbf{\#\#\# Input:}\\
\uline{Please tell me how to get to Osaka from Tokyo.}~\footnotemark[\value{footnotecommon1}]\\
\textbf{\#\#\# Output (Stormy):}\\
\uline{Depart from Tokyo by plane. When you arrive in Osaka, transfer to the Shinkansen and go to Nagoya. From Nagoya, take the Kintetsu Limited Express.}~\footnotemark[\value{footnotecommon1}]\\
\textbf{\#\#\# Output (CALM):}\\
\uline{Which is better, train or Shinkansen?\\
* ``Train'' refers to all types of transportation, such as JR trains and private railways.\\
* ``Shinkansen'' refers to the Tokaido and Sanyo Shinkansen that run between Tokyo and Shin-Osaka.\\
* ``Airplane'' typically refers to the route between Haneda and Itami.\\
* ``Bus'' refers to the Toei buses that operate in Tokyo.\\
* ``Car'' includes not only private cars but also taxis and rental cars.\\
* ``Bicycle'' includes electric-assisted bicycles.\\
* ``Walking,'' ``jogging,'' and ``strolling'' are excluded.}~\footnotemark[\value{footnotecommon1}]\\
\textbf{\#\#\# Output (Instruct LLaMA 7B):}\\
\uline{Please tell me how to get to Tokyo from Osaka.}~\footnotemark[\value{footnotecommon1}]\\
\textbf{\#\#\# Output (LLaMA 7B):}\\
\uline{Please tell me how to get to Osaka from Tokyo.}~\footnotemark[\value{footnotecommon1}]
\end{brekableitembox}

\vspace{-2mm}\begin{brekableitembox}{Comparison of Responses (3)}
\fontsize{9.5pt}{13pt}\selectfont\textbf{\#\#\# Input:}\\
\uline{Where is the University of Tokyo?}~\footnotemark[\value{footnotecommon1}]\\
\textbf{\#\#\# Output (Stormy):}\\
\uline{The University of Tokyo is a national university located in Bunkyo-ku, Tokyo.}~\footnotemark[\value{footnotecommon1}]\\
\textbf{\#\#\# Output (CALM):}\\
\uline{The University of Tokyo is located at 7-3-1 Hongo, Bunkyo-ku, Tokyo. The latitude and longitude of the University of Tokyo are 35.676044, 135.869422.\\
\\\\
~~\#\#Answer: What do you use to get to that address? (Maps or Google Map, etc.)\\
\\
~~\#\#Explanation: What mode of transportation should I use to get to that address?\\
\\
~~\#\#Answer: Which is better to use, train or bus?\\
\\
~~\#\#Explanation: If you use the JR line, please get off at the ``Yamanote Line.'' The JR line goes to ``Tokyo'' station.\\
\\
~~\#\#Answer: Where should I go from Tokyo station?\\
\\
~~\#\#Explanation: The trains you take from ``Tokyo Station'' are the ``Tokaido Shinkansen,'' ``Tohoku, Joetsu, Hokuriku Shinkansen'' and ``Limited Express Hitachi.''\\
\\
~~\#\#Answer: Which is faster and cheaper, the Joban Line or the Shinkansen?\\
\\
~~\#\#Explanation: The Shinkansen is faster. However, the price is different.\\
\\
~~\#\#Answer: I don't have money, what should I do?\\
\\
~~\#\#Explanation: First, please prepare money. If you have a credit card, you can use it.\\
\\
~~\#\#Answer: How should I pay?\\
\\
~~\#\#Explanation: You can pay by card. VISA card}~\footnotemark[\value{footnotecommon1}]\\
\textbf{\#\#\# Output (Instruct LLaMA 7B):}\\
\uline{The University of Tokyo is in Tokyo.}~\footnotemark[\value{footnotecommon1}]\\
\textbf{\#\#\# Output (LLaMA 7B):}\\
\uline{Where is the University of Tokyo?}~\footnotemark[\value{footnotecommon1}]\\
(Authors' note: The correct address is 7-3-1 Hongo, Bunkyo-ku, Tokyo.)
\end{brekableitembox}

\vspace{-2mm}\begin{brekableitembox}{Comparison of Responses (4)}
\fontsize{9.5pt}{13pt}\selectfont\textbf{\#\#\# Input:}\\
\uline{The phone is ringing. What should I do?}~\footnotemark[\value{footnotecommon1}]\\
\textbf{\#\#\# Output (Stormy):}\\
\uline{Should you answer the phone? Or should you put it on hold?}~\footnotemark[\value{footnotecommon1}]\\
\textbf{\#\#\# Output (CALM):}\\
\uline{Press the hold button, answer the phone, or call the answering service. When you hear the ringtone, reply within 10 seconds. If there is no response, try calling again. If that still doesn't work, proceed to the next step.\\
* Press `0' to end the call.\\
* Press `*' after the ringtone. This will enable the call.}~\footnotemark[\value{footnotecommon1}]\\
\textbf{\#\#\# Output (Instruct LLaMA 7B):}\\
\uline{The phone is ringing. What should I do?}~\footnotemark[\value{footnotecommon1}]\\
\textbf{\#\#\# Output (LLaMA 7B):}\\
\uline{The phone is ringing. What should I do?}~\footnotemark[\value{footnotecommon1}]
\end{brekableitembox}

\vspace{-2mm}\begin{brekableitembox}{Comparison of Responses (5)}
\fontsize{9.5pt}{13pt}\selectfont\textbf{\#\#\# Input:}\\
\uline{What are the three major festivals in Kyoto?}~\footnotemark[\value{footnotecommon1}]\\
\textbf{\#\#\# Output (Stormy):}\\
\uline{Aoi Festival, Gion Festival, Jidai Festival}~\footnotemark[\value{footnotecommon1}]\\
\textbf{\#\#\# Output (CALM):}\\
\uline{The three are Gion Festival, Aoi Festival, and Jidai Festival. They are also referred to as the three major festivals of Kyoto.}~\footnotemark[\value{footnotecommon1}]\\
\\\\
~~\#N/A\\
~~\#N/A\\
(Authors' Note: ``~~\#N/A'' continues from here.)\\
\textbf{\#\#\# Output (Instruct LLaMA 7B):}\\
\uline{What are the three major festivals in Kyoto?}~\footnotemark[\value{footnotecommon1}]\\
\textbf{\#\#\# Output (LLaMA 7B):}\\
\uline{What are the three major festivals in Kyoto?}~\footnotemark[\value{footnotecommon1}]\\
(Authors' note: The correct answer is Aoi, Gion, Jidai festivals.)
\end{brekableitembox}

Stormy, which was instruct-tuned from CALM, demonstrates a more natural response format.
On the other hand, the difference between Instruct LLaMA 7B and LLaMA 7B appears to be minimal, with only a slight improvement in the accuracy of response comparison (3).
According to the literature~\cite{hirano2023llmjapanesedataset}, the small improvement in LLaMA 13B's 1 epoch tuning was attributed to the limited amount of training.
However, despite 5 epochs of tuning on LLaMA 7B, Instruct LLaMA 7B did not show significant improvement.
This result contrasts with the trend observed in the quantitative evaluations discussed in Section \ref{sec:quantitative-eval}.
This difference suggests that instruction tuning alone may not lead to significant improvements.
Not only through instruction tuning, but also by additional pre-training to accumulate knowledge about the target language (in this case, Japanese) and then performing instruction tuning, there is a potential to improve performance in the target language~\cite{yong2023bloom1}.
Various methods can be considered for pre-training, including pre-training from scratch solely in Japanese or conducting additional pre-training in Japanese using English or multilingual models.
Hence, exploring methods to achieve high performance in languages other than English will be a future challenge.
Moreover, the difference between the trend observed in the qualitative evaluation and that of the quantitative evaluation highlights the importance of evaluating qualitative performance not only based on simple metrics like likelihood or perplexity generated by the model but also based on the actual output strings.

\section{Related Work}
\subsection{Pre-training Models}
The Transformer~\cite{Vaswani2017}, a crucial component of large language models (LLMs), consists of two architectures: encoder and decoder.
Models such as BERT~\cite{Devlin2018}, RoBERTa~\cite{Liu2019roberta}, and DeBERTa~\cite{he2021deberta,he2023debertav3} utilize the Transformer's encoder.
In recent LLMs, the decoder (causal decoder) is primarily used.
The GPT series~\cite{GPT-1,GPT-2,GPT-3,gpt4} is a representative language model with a decoder architecture.
There are also many proposed models, such as OPT~\cite{zhang2022opt}, GPT-NeoX-20B~\cite{black-etal-2022-gptneox}, Gopher~\cite{rae2022gopher}, PaLM~\cite{chowdhery2022palm}, BLOOM~\cite{scao2022bloom}, Pythia~\cite{biderman2023pythia}, and LLaMA series~\cite{touvron2023llama,touvron2023llama2}.
Although relatively few, there is also T5~\cite{2020t5} as an encoder-decoder architecture, and Flan T5~\cite{wei2022flan} has been proposed as an extension of it.

\subsection{Tuning after Pre-training}
Through pre-training, LLMs acquire the ability to solve various tasks.
However, there is an increasing number of cases where fine-tuning is performed to align more with specific purposes, such as dialogue responses.
There are mainly two approaches for this fine-tuning: alignment tuning and instruction tuning.
Alignment tuning is a tuning to aim for outputs more in line with human preferences~\cite{ouyang2022instructgpt}.
Through alignment tuning, LLMs are adjusted to produce outputs that align with human values, such as being helpful, honest, and harmless.
In recent LLMs, reinforcement learning (RL)~\cite{christiano2017rlhf}, especially Proximal Policy Optimization (PPO)~\cite{schulman2017ppo}, is used to learn from human-labeled response preference rankings.
Instruction tuning is generally a method of fine-tuning LLMs with datasets in natural language format, supervised~\cite{ouyang2022instructgpt}, and performed using various tasks (multi-task)~\cite{sanh2022multitaskp3}.
It has been shown that instruction tuning can demonstrate excellent performance even for unknown tasks~\cite{wei2022flan,chung2022scaling,sanh2022multitaskp3}.
In instruction tuning, instructions that request the output of the LLM are described in the input, and training is performed so that the LLM outputs the expected response.
It has been shown that the task-explanation part is particularly decisive in the performance of the LLM that improves through instruct tuning~\cite{wei2022flan}.
If tuning is performed with a dataset that is closer to traditional supervised learning by removing the explanation about the task, the performance will significantly decrease compared to not removing it.
Most cases of instruct tuning are performed with formatted existing datasets for various tasks in natural language format for instruction tuning datasets.
Specifically, labeled datasets are applied with instructions written by humans explaining the task, explaining the direction of the output, and instructing the LLM to understand the task from the input~\cite{sanh2022multitaskp3,wei2022flan,wang-etal-2022-super}.
Other construction methods include examples of constructing datasets using the output of ChatGPT or GPT-4~\cite{codealpaca,peng2023instruction,alpaca}\footnote{\url{https://huggingface.co/datasets/RyokoAI/ShareGPT52K}}$^{,}$\footnote{\url{https://github.com/teknium1/GPTeacher}}, and there are few examples of constructing datasets manually~\cite{dolly}.

\subsection{Tuning of LLMs}
Efficient tuning in LLMs with many parameters is attracting attention to adapt LLMs to various downstream tasks.
In particular, LoRA~\cite{hu2022lora} is widely applied to open-source LLMs.
For example, Alpaca-LoRA~\cite{alpaca-lora} uses LoRA to tune LLaMA 7B as a lightweight tuning version of Alpaca~\cite{alpaca}.
Also, AdaLoRA~\cite{zhang2023adalora} changes the value of the rank in LoRA.
This adjustment occurs according to the layer to be applied.

Other efficient tuning methods include adding an Adapter layer to the existing layers~\cite{houlsby2019parameterefficient,lin-etal-2020-exploring,hu2023llmadapters}, and prompt tuning~\cite{li-liang-2021-prefix,lester-etal-2021-power}, which fixes the weights of the pre-trained model, adds trainable parameters to the prompt instructions.

\section{Conclusion}
We constructed an instruction dataset for Japanese-based LLMs (Large Language Models).
The dataset excludes any translation data originally present in the llm-japanese-dataset and introduces additional tasks to the existing ones.
We performed LoRA tuning on LLMs pre-trained in both Japanese and English, respectively.
The tuning was done using Japanese instruction data.
We evaluated the tuned models from both quantitative and qualitative perspectives.
The results show that tuning with Japanese instruction data improves performance in quantitative evaluations.
In particular, the results indicate that not only Japanese-based models but also English-based models can be tuned in Japanese using the Japanese instruction dataset.
Furthermore, even with smaller model sizes like 7B or 13B, instruction tuning can sometimes improve performance in downstream tasks, suggesting a result different from prior research. 

Future research can address not only comparing the likelihood of the current model's output, but also using the actual output in the evaluation of the model.
Additionally, it could include evaluation from the perspective of human preference in Japanese.

\section*{Acknowledgment}
This work was supported in part by JSPS KAKENHI Grant Number JP21K12010 and JST PRESTO Grant Number JPMJPR2267.

\bibliographystyle{IEEEtran}
\bibliography{myrefs}

\appendix{Prompts Used in JNLI and MARC-ja}\label{sec:prompt-jnli-marcja}
\begin{brekableitembox}{Example of 1-shot prompt used in JNLI (v0.2)}
\fontsize{9.5pt}{13pt}\selectfont\uline{Please answer the relationship between the premise and the hypothesis from}\footnotemark[\value{footnotecommon1}] entailment, contradiction, \uline{and}\footnotemark[\value{footnotecommon1}] neutral.\\\\
\uline{Constraints:\\
- If the hypothesis can be derived from the premise using logical or common sense knowledge, output}\footnotemark[\value{footnotecommon1}] entailment\\
\uline{- If the premise and the hypothesis are incompatible, output}\footnotemark[\value{footnotecommon1}] contradiction\\
\uline{- If neither of the above, output}\footnotemark[\value{footnotecommon1}] neutral\\\\
\uline{Premise: Two women are jumping to catch a frisbee in the grass.\\
Hypothesis: The two women are holding a tray with donuts on it.\\
Relationship:}\footnotemark[\value{footnotecommon1}] entailment\\
\\
\uline{Premise: There are two children, and bananas and kiwis are placed next to the mixer.\\
Hypothesis: There are children with droppers at the table where the mixer is placed.\\
Relationship:}~\footnotemark[\value{footnotecommon1}]
\end{brekableitembox}

\setcounter{footnote}{9}
\newcounter{footnotecommon2}
\footnotetext{Although the same instructions are repeated within the same prompt, the template from the implementation reference (\url{https://github.com/Stability-AI/lm-evaluation-harness/tree/jp-stable}) is used as is.}\setcounter{footnotecommon2}{\value{footnote}}
\vspace{-2mm}\begin{brekableitembox}{Example of 1-shot prompt used in JNLI (v0.3)~\footnotemark[\value{footnotecommon2}]}
\fontsize{9.5pt}{13pt}\selectfont\uline{Below is a combination of instructions explaining the task and contextual inputs. Write a response that adequately meets the request.\\\\\\
\#\#\# Instructions:\\
Please answer the relationship between the given premise and hypothesis.\\
\\
Choose your output from the following:}~\footnotemark[\value{footnotecommon1}]\\
entailment\\
contradiction\\
neural\\
\\
\uline{\#\#\# Input:\\
Premise: Two women are jumping to catch a frisbee in the grass.\\
Hypothesis: The women are trying to catch a frisbee.\\
\\
\#\#\# Response:}~\footnotemark[\value{footnotecommon1}]\\
entailment\\
\\
\uline{\#\#\# Instructions:\\
Please answer the relationship between the given premise and hypothesis.\\
\\
Choose your output from the following:}~\footnotemark[\value{footnotecommon1}]\\
entailment\\
contradiction\\
neural\\
\\
\uline{\#\#\# Input:\\
Premise: There are two children, and bananas and kiwis are placed next to the mixer.\\
Hypothesis: There are children with droppers at the table where the mixer is placed.\\
\\
\#\#\# Response:
\\\,}~\footnotemark[\value{footnotecommon1}]
\end{brekableitembox}

\vspace{-2mm}\begin{brekableitembox}{Example of 1-shot prompt used in MARC-ja (v0.2)}
\fontsize{9.5pt}{13pt}\selectfont\uline{Please classify the product review into either} negative \uline{or} positive \uline{sentiment. Please lowercase the output.}~\footnotemark[\value{footnotecommon1}]\\
\\
\uline{Product Review: I like country and initially thought of buying the CD, the movie has a decent story, it's okay\\
Sentiment:}\footnotemark[\value{footnotecommon1}] positive\\
\\
\uline{Product Review: I enjoyed it till the end. Personally, I wanted to see more dance scenes. I hope it will be staged.\\
Sentiment:}~\footnotemark[\value{footnotecommon1}]
\end{brekableitembox}

\vspace{-2mm}\begin{brekableitembox}{Example of 1-shot prompt used in MARC-ja (v0.3)~\footnotemark[\value{footnotecommon2}]}
\fontsize{9.5pt}{13pt}\selectfont\uline{Below is a combination of instructions explaining the task and contextual inputs. Write a response that adequately meets the request.\\
\\\\
\#\#\# Instructions:\\
Please classify the following product review into either a positive or negative sentiment class.\\
\\\\
\#\#\# Input:\\
I like country and initially thought of buying the CD, the movie has a decent story, it's okay\\
\\
\#\#\# Response:\\
Positive\\
\\
\#\#\# Instructions:\\
Please classify the following product review into either a positive or negative sentiment class.\\
\\\\
\#\#\# Input:\\
I enjoyed it till the end. Personally, I wanted to see more dance scenes. I hope it will be staged.\\
\\
\#\#\# Response:
\\\,}~\footnotemark[\value{footnotecommon1}]
\end{brekableitembox}

\end{document}